\documentclass[]{bytedance}
\usepackage[toc,page,header]{appendix}

\usepackage{amsmath}
\usepackage{soul}
\usepackage{multirow}
\definecolor{topcolor}{RGB}{130,195,235}
\definecolor{secondcolor}{RGB}{220,240,250}
\usepackage[table]{xcolor}
\definecolor{sectiongray}{RGB}{220,220,220}
\usepackage{makecell}
\usepackage{amssymb}
\usepackage{multirow}
\usepackage{arydshln}
\usepackage{newfloat}
\usepackage{listings}
\usepackage{booktabs}
\usepackage{minitoc}
\usepackage{amsfonts}
\usepackage{amssymb}
\usepackage{tabularx}
\usepackage{listings}
\usepackage{xcolor}
\usepackage{cancel}

\usepackage{tabulary,multirow,xspace}
\usepackage{fixmath,mathtools,nicefrac,mmstyle}
\usepackage{subcaption}
\usepackage{caption}
\usepackage{wrapfig} 
\usepackage[misc]{ifsym} 
\usepackage{colortbl}
\usepackage{wrapfig}
\usepackage{multicol}
\usepackage[most]{tcolorbox}
\usepackage{pifont}
\usepackage{pifont}
\usepackage{xcolor}
\usepackage{listings}

\newcommand{\name}{DART-SD}
\title{\name{}: Diamond-topology Aware Retrieval and Tuning for Self-Distillation of \\ Multi-Turn Tool-Calling Agents}

\author[1,\star]{Hangrui Xu}
\author[1,\star]{Jiarui Wang}
\author[1,\star,\dagger]{Yang Yang}
\author[1]{Chuanbo Zhu}
\author[1]{Fangda Chen}
\author[1]{\\ Ziqi Wu}
\author[1]{Jingming Cai}
\author[2,\dagger]{Yan Song}

\affiliation[1]{ByteDance}
\affiliation[2]{University of Science and Technology of China}
\contribution[\star]{Equal contribution}
\contribution[\dagger]{Corresponding author}

\abstract{
Equipping Large Language Models (LLMs) with multi-turn tool-calling capabilities is essential for building autonomous agents. However, progress is fundamentally limited by the reliance on full-length trajectory imitation. For tasks involving multiple order-independent sub-goals, the optimal solution space forms a vast combinatorial diamond lattice. Forcing this rich topology into monolithic trajectories causes a severe \textit{topological collapse}, indiscriminately penalizing valid alternative explorations and severely degrading policy diversity. To address this, we propose \textbf{DART-SD} (\textit{\underline{\textbf{D}}iamond-topology \underline{\textbf{A}}ware \underline{\textbf{R}}etrieval and \underline{\textbf{T}}uning for \underline{\textbf{S}}elf-\underline{\textbf{D}}istillation}), a novel framework that shifts the paradigm from global forcing to topology-guided localized correction. 
DART-SD first models the execution process as a converging \textit{Interaction-State Transition Graph} (\textbf{ISTG}), faithfully capturing the inherent diamond topology of successful and failed exploratory paths. During autonomous rollouts, the framework identifies the \textit{Critical Topological Breakpoint} (\textbf{CTB}) and retrieves success-supported recovery references. Finally, we introduce a \textit{progressive self-distillation paradigm} through CTB-guided localized supervision, ensuring that the training loss is calculated exclusively on the generated recovery steps while strictly protecting the valid reasoning prefix from destructive gradient updates. Experiments on complex multi-turn tool-calling benchmarks demonstrate that DART-SD significantly outperforms traditional full-trajectory baselines.
}

\correspondence{Yang Yang at \email{yang.yves@bytedance.com}, Yan Song at \email{clksong@gmail.com}}

\begin{document}
\maketitle

\section{Introduction}

Equipping Large Language Models (LLMs) with multi-turn tool-calling capabilities is an essential milestone toward building autonomous agents capable of solving complex, real-world tasks \cite{xi2025rise, qin2025tool, wu2026models}. Through sequential interactions with external environments, tool-augmented agents can retrieve up-to-date information, execute code, and manipulate APIs to transcend their static parametric knowledge \cite{chen2026learning, yang2026reasoningreinforcementlearningunlocks, gekhman2026thinkingrecallreasoningunlocks, wu2026promsa, r3g}. Currently, state-of-the-art tool-calling performance is predominantly achieved by massive and expensive frontier models \cite{google2025gemini, anthropic2025claude4}. Consequently, agent distillation has gained significant research momentum, aiming to transfer these sophisticated planning and interaction capabilities into efficient, compact open-source models \cite{yang2026reasoningreinforcementlearningunlocks, ye2025tltraining, kang2026distilling}.

\begin{figure}
    \centering
    \includegraphics[width=0.8\linewidth]{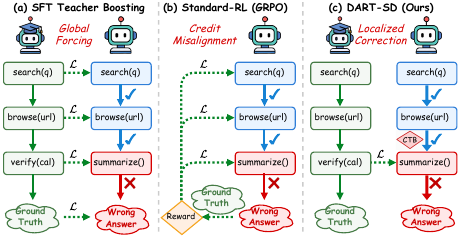}
    \caption{
    Comparison of training paradigms. (a) SFT Teacher Boosting applies an indiscriminate global loss, which overwrites valid exploration. (b) Standard-RL (GRPO) misassigns credit through a uniformly distributed reward spread. (c) DART-SD dynamically identifies the Critical Topological Breakpoints (CTB) and applies localized correction while preserving valid exploration.
    }
    \vspace{-0mm}
    \label{fig:teaser}
\end{figure}

The prevalent agent distillation paradigm relies on full trajectory imitation via Behavior Cloning (BC) or Supervised Fine-Tuning (SFT) \cite{qin2024toolllm}. In long-horizon tasks,  global forcing paradigms apply an indiscriminate global loss that overwrites the student model's valid exploratory steps \cite{meng2026sparsecriticaltokenlevelanalysis}, causing models to memorize redundant trajectories instead of extracting the task's core logical backbone. 
Recent efforts attempt to explore hindsight-based scaffolding or Reinforcement Learning (RL) \cite{ye2025feedback, opsd2026, hhd2025}. However, traditional hindsight methods \cite{hint2026} still treat multi-turn interactions as strict linear sequences. This rigid linearity forces models to confuse fatal errors with harmless exploration, severely degrading token efficiency and internal knowledge consistency \cite{mousavi2026doeslossoptimizationactually, zheng2024reliablellmsknowledgebases}. Meanwhile, standard RL approaches \cite{shao2024deepseekmath,wang2026group}, such as Group Relative Policy Optimization (GRPO), frequently suffer from credit misassignment due to uniformly distributed rewards spread across all intermediate tool calls and inadvertently penalize valid steps within failed trajectories (see Figure \ref{fig:teaser}).

The monolithic SFT and standard RL paradigms are fundamentally limited by misrepresenting the multi-turn tool-calling process as a collection of isolated linear trajectories \cite{chai2025parl, hu2025training}. This oversimplified perspective ignores the inherent graph structure of state transitions \cite{gupta2026world, kausik2026context}. For tasks with order-independent sub-goals, distinct trajectories frequently intersect at shared intermediate states, structuring the optimal solution space into a vast combinatorial diamond lattice \cite{gallart2026chain}. Forcing models to flatten this interconnected topological space into single linear paths, traditional methods inevitably cause a severe \textit{topological collapse}. This structural oversight induces myopic credit assignment \cite{wang2026group}, arbitrarily penalizing valid alternative explorations and severely degrading policy diversity \cite{lightman2024let, yu2025building}.

To align with this topological reality, we propose \textbf{DART-SD} (\textit{\textbf{D}iamond-topology \textbf{A}ware \textbf{R}etrieval and \textbf{T}uning for \textbf{S}elf-\textbf{D}istillation}), a framework that shifts the paradigm from linear imitation to topology-aware localized correction. DART-SD extracts a structural prior from teacher rollouts, instantiated as an \textit{Interaction-State Transition Graph} (\textbf{ISTG}) covering both successful and failed trajectories. By defining nodes as cumulative interaction states rather than transient actions, the ISTG captures the diamond topology, naturally resolving order-dependency conflicts. During self-distillation, DART-SD projects student states onto the success-reachable region to identify the \textit{Critical Topological Breakpoint} (\textbf{CTB}) and its recovery anchors. From each anchor, it searches the ISTG for a success-supported continuation as a reference for recovery generation. Finally, DART-SD applies a \textit{progressive self-distillation paradigm} through CTB-guided localized supervision, ensuring that the loss is calculated only on recovery steps to protect valid reasoning prefixes.
 
In summary, our main contributions are as follows:

\begin{itemize}
\item We identify a fundamental topological collapse in existing agent-training paradigms. To resolve this, we introduce the Interaction-State Transition Graph (\textbf{ISTG}), which represents tool execution through cumulative interaction states and captures the diamond topology induced by order-independent exploration.

    \item We define the \textit{Critical Topological Breakpoint} (\textbf{CTB}) by projecting student interaction states onto the empirical success-reachable region. We propose \textit{CTB-guided localized supervision}, which retrieves success-supported references and supervises the generated recovery steps while strictly preserving valid prefixes.

    \item We propose a \textit{progressive self-distillation paradigm} that repeatedly
rolls out the student, identifies its evolving capability
boundary, and performs CTB-guided localized supervision.
This iterative process enables the student to continuously
extend valid interaction prefixes and progressively master more complex
tool-use behaviors.
    
    \item Extensive experiments across five benchmarks and two model scales demonstrate that DART-SD consistently outperforms supervised and reinforcement-learning baselines, while reducing redundant tool calls and extending valid interaction prefixes.
\end{itemize}

\section{Related Work}

\textbf{Distillation-Based Paradigms.}
Early attempts at building autonomous agents 
rely on Behavior Cloning (BC) or Supervised Fine-Tuning (SFT) over complete expert trajectories \cite{qin2024toolllm}. However, this global forcing paradigm forces students to mimic entire sequences, often leading to compounding errors in long-horizon tasks as deviations propagate autoregressively \cite{score2026}. Furthermore, indiscriminate token-level imitation triggers exposure bias \cite{ranzato2015sequence} and overwrites the student’s valid exploration habits. To address the limitations of static global forcing, recent distillation methods have shifted toward student-centered and dynamic strategies \cite{liao2026multi}. SCoRe-SFT \cite{score2026} provides post-hoc scaffolding through student-centered knowledge distillation, while On-Policy Self-Distillation (OPSD) \cite{opsd2026} mitigates offline distribution shifts by dynamically rolling out and distilling trajectories from the current student policy. Similarly, hindsight-based distillation methods such as HINT-SD \cite{hint2026} leverage targeted scaffolding for correction.

\noindent
\textbf{Reinforcement Learning Paradigms.}
To further refine policy optimization beyond distillation, Reinforcement Learning (RL) approaches leverage environment feedback to directly align agent behaviors. Methods such as FTRL-GRPO \cite{ye2025feedback} and ToolRL \cite{qian2026toolrl} optimize agents using sparse outcome rewards and verifiable execution feedback. However, sparse terminal signals in these methods often lead to severe credit misassignment, inadvertently penalizing valid intermediate steps within failed trajectories \cite{wang2025information, zeng2025reinforcing, cheng2026beyond, shao2024deepseekmath, ye2025feedback, meng2026sparsecriticaltokenlevelanalysis}. Even finer-grained RL methods such as MatchTIR \cite{qu2026matchtir}, which aligns trajectories via bipartite matching to provide turn-level rewards, still fundamentally treat multi-turn interactions as rigid linear sequences, thereby overlooking the inherent topology of exploration.  In contrast, our proposed DART-SD shifts the paradigm from linear imitation and complex matching to topology-aware localized correction. By calculating the optimization loss exclusively on retrieved recovery steps, DART-SD strictly protects valid reasoning prefixes from destructive gradient updates.

\section{Methodology}
\label{sec:method}

\begin{figure*}[t]
    \centering
    \includegraphics[width=\textwidth]{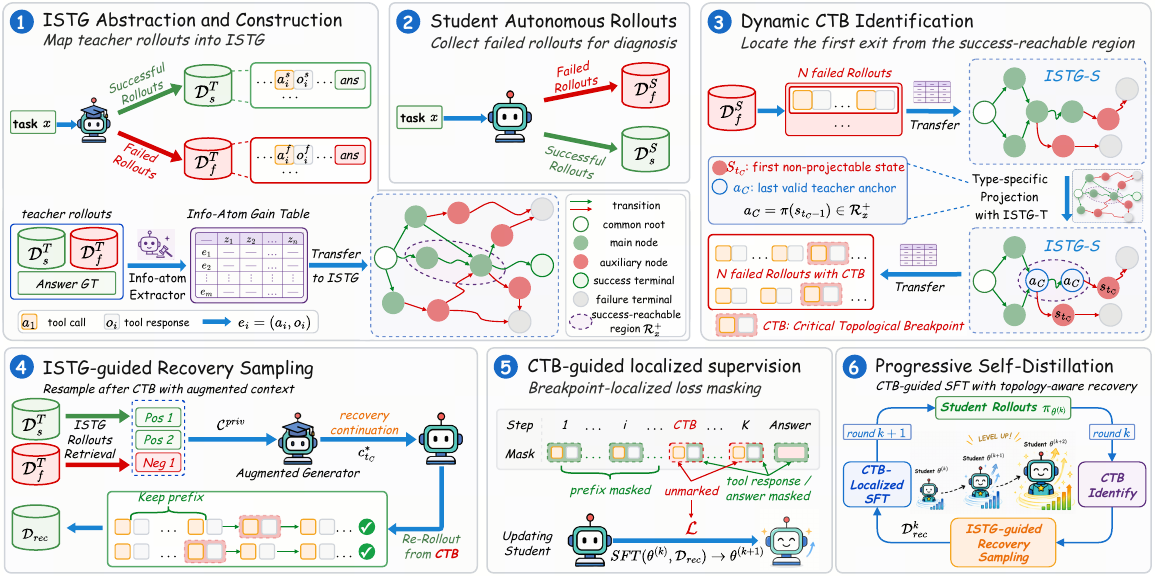}
\caption{
Overview of DART-SD.
(1) DART-SD constructs an ISTG from teacher rollouts, where main and auxiliary nodes model information acquisition and useless exploration.
(2--3) Failed student rollouts are collected and replayed in the same interaction-state space, then projected onto the budget-filtered success-reachable region \(\mathcal{R}_x^+\), where the first projectable-to-non-projectable transition defines the CTB.
(4--5) Conditioned on the retained student prefix and privileged teacher references, DART-SD generates a recovery continuation after the CTB and applies localized supervision only to the generated assistant tokens.
(6) This CTB-guided SFT loop is progressively repeated across self-distillation rounds.
}
    \label{fig:method}
\end{figure*}

Figure~\ref{fig:method} provides an overview of the DART-SD framework. DART-SD first constructs an ISTG from teacher rollouts, then identifies the CTB of failed student rollouts through success-reachable projection. The identified CTB is used to generate localized recovery continuations and to drive progressive self-distillation.

\subsection{ISTG Abstraction and  Construction}
\label{sec:interaction_graph}
\textbf{Information Atom Abstraction.} For each task $x$, useful facts extracted from tool responses are normalized into a task-specific set of information atoms $\mathcal{K}_x$. Semantically equivalent responses share the same atom, while non-informative responses contribute none. We realize this in two stages. A deterministic stage canonicalizes each tool response and decides whether it carries task-usable data at all. Parsing the response into fields, it is treated as non-informative only when every field is either a status signal or an empty or placeholder value, so a payload retaining a single substantive value still qualifies as informative, whereas an error or not-found message does not. Non-informative responses are collapsed per tool into a single class, which removes the bulk of the corpus without semantic judgement.

A semantic stage then assigns atoms to the surviving candidates. Let $e$ denote a tool call, with tool identity $\operatorname{tl}(e)$ and canonicalized response $\bar{o}(e)$. All candidates of a task are examined jointly, conditioned on the task question and one successful rollout, yielding a set-valued atom map
\begin{equation}
    \alpha_x:\ \bigl(\operatorname{tl}(e),\bar{o}(e)\bigr)\ \longmapsto\ \alpha_x(e)\subseteq\mathcal{K}_x,
    \qquad
    \lvert\alpha_x(e)\rvert\leq 1 .
    \label{eq:atom_map}
\end{equation}
An informative response is mapped to the singleton containing its atom, while a non-informative response, or one left unjudged, is mapped to $\varnothing$. Responses supplying the same fact are constrained to receive the same atom, so each element of $\mathcal{K}_x$ is the canonical representative of one class of information-equivalent calls, namely those whose images under $\alpha_x$ coincide and are non-empty. Judging a task's candidates jointly is what makes this equivalence decidable: atoms are assigned with all alternatives in view, rather than labeling responses independently and relying on the labels to agree.

The map $\alpha_x$ supplies the acquisition increment used in Eq.~\eqref{eq:node_type}. Writing $B_t$ for the tool call or concurrent call bundle executed at step $t$,
\begin{equation}
    \Delta I_t
    =
    \left\{
        k \in \mathcal{K}_x
        \;\middle|\;
        \exists e\in B_t,\;
        k\in\alpha_x(e),\;
        k\notin I_{t-1}
    \right\},
    \qquad
    I_t = I_{t-1}\cup\Delta I_t .
    \label{eq:atom_gain}
\end{equation}
Two properties of this construction determine the resulting topology. First, $\alpha_x$ identifies responses across surface form and across tools, so the same fact returned as free text by one tool and as a structured record by another yields one atom; the two acquisition paths therefore reconverge at the same main information state instead of forking into states the projection would treat as unrelated. Second, mapping non-informative and unjudged responses to
$\varnothing$ is what makes them contribute none: such a call
contributes no element to $\Delta I_t$ in
Eq.~\eqref{eq:atom_gain} and leaves the information set unchanged,
instead of receiving an identifier derived from its own text, which
would manufacture a distinct information state that no evidence
supports and inflate the information dimension of the graph. Because only informative responses yield atoms, residual judgement errors predominantly omit an atom rather than invent one, leaving the graph sparser but never fabricating states or reachability.

\textbf{Main and Auxiliary Nodes.}
For a task \(x\), the interaction state at step \(t\) is represented as
\begin{equation}
    X_t=(I_t,U_t),
    \label{eq:interaction_state}
\end{equation}
where \(I_t\) is the set of canonical information atoms acquired up to
step \(t\), and \(U_t\) records the multiset of useless operations performed since
the most recent main node.

The root state \(X_0=(\varnothing,\varnothing)\) is defined as a main
node. For \(t\geq 1\), the type of \(X_t\) is determined by the
preceding transition:
\begin{equation}
    \operatorname{type}(X_t)=
     \begin{cases}
        \mathrm{main}, & \Delta I_t\neq\varnothing,\\
        \mathrm{aux}, & \Delta I_t=\varnothing
        \ \text{and}\ U_t\neq\varnothing.
    \end{cases}
    \label{eq:node_type}
\end{equation}
where \(\Delta I_t\) is the set of previously unseen information atoms
acquired at step \(t\), given by Eq.~\eqref{eq:atom_gain}.

Main nodes include mixed steps acquiring new information alongside useless operations and are represented with \(U_t=\varnothing\), whereas auxiliary nodes record useless operations after the main node without changing the information set.

\textbf{Interaction-State Transition Graph (ISTG).}
For each task \(x\), we construct a directed multigraph
\begin{equation}
    G_x=(V_x,E_x).
    \label{eq:istg}
\end{equation}
The graph is built from all successful and failed teacher rollouts.
All rollouts share a common root and terminate at either a success or a
failure terminal. Each vertex \(v\in V_x\) represents a state
\(X(v)=(I(v),U(v))\) with type
\(\operatorname{type}(v)\in\{\mathrm{main},\mathrm{aux}\}\), and each
edge represents one tool call or concurrent call bundle. Parallel edges
are retained.

Main nodes form the information-acquisition backbone, while auxiliary
nodes represent useless exploration attached to the most recent main
node. Both node types and transitions are retained, since
auxiliary exploration contributes to execution depth and matching.
The set-valued information component allows order-independent
acquisition paths to diverge and reconverge at the same main
information state, producing the diamond structures of
the ISTG.
A student rollout is replayed under the same state-update rule:
\begin{equation}
    P_x^s=(s_0,\ldots,s_T),
    \qquad
    X(s_t)=(I_t^s,U_t^s).
    \label{eq:student_path}
\end{equation}

The node type of \(s_t\) is determined by the preceding transition.
Teacher and student executions are therefore embedded in a shared
interaction-state space, enabling separate matching of information
acquisition and useless exploration without requiring action-wise
correspondence.

\subsection{Success-Reachable Projection and CTB}
\label{sec:topological_breakpoints}
\textbf{Empirical Success-Reachable Region.}
Let \(\mathcal{T}_x^+\) denote the successful teacher rollouts for task
\(x\), and let \(d_x^{\min}\) be the minimum number of graph transitions
from the root to a success terminal among these rollouts. We define the
task-specific reachability budget as
\begin{equation}
B_x
=
\min\bigl(d_x^{\min}+\Delta_x,\; B_x^{\max}\bigr),
\end{equation}
where \(\Delta_x\) is the task-dependent allowance beyond the shortest
successful depth and \(B_x^{\max}\) is the budget cap.
For a node \(v\) on a successful rollout \(\tau\), let
\(r_\tau(v)\) denote the number of graph transitions remaining from \(v\)
to the success terminal of \(\tau\). The empirical success-reachable
region is defined by
\begin{equation}
\mathcal{R}_x^+
=
\left\{
v\in V_x \;\middle|\;
\exists\,\tau\in\mathcal{T}_x^+,\;
v\in\tau,\;
r_\tau(v)\le B_x
\right\}.
\end{equation}
Thus, \(\mathcal{R}_x^+\) contains both main and auxiliary nodes on
successful teacher rollouts whose remaining distance to success is at
most \(B_x\). The distance is measured backward from the success
terminal along the rollout, rather than as forward depth from the root.
States observed exclusively on failed rollouts remain in \(G_x\), but are
excluded from \(\mathcal{R}_x^+\) and cannot serve as recovery anchors.

\textbf{Type-specific State Projection.}
For each student state \(s_t=(I_t^s,U_t^s)\), we further apply the
projection rule associated with its node type. Let
\(\mathcal{R}_{x,\mathrm{main}}^+\) and
\(\mathcal{R}_{x,\mathrm{aux}}^+\) denote the main and auxiliary nodes in
the budget-filtered empirical success-reachable region.

For a student main node, the acceptable teacher anchors are defined over non-root successful nodes with valid information acquisition:
\begin{equation}
\mathcal{A}_t^{\mathrm{main}}
=
\left\{
v\in\mathcal{R}_{x,\mathrm{main}}^+
\;\middle|\;
I(v)\subseteq I_t^s
\right\}.
\end{equation}
Thus, a student main node can be aligned with any reachable teacher main
node whose information set is contained in the student's acquired
information. Equality of information sets and the useless-operation
component are not required. For mixed transitions, the criterion depends
only on the information component, since any transition with information
gain is classified as a main transition.

For a student auxiliary node, we first consider the reachable teacher
main nodes whose information is contained in \(I_t^s\):
\begin{equation}
\mathcal{M}_t
=
\left\{
m\in\mathcal{R}_{x,\mathrm{main}}^+
\;\middle|\;
I(m)\subseteq I_t^s
\right\}.
\end{equation}
If \(\mathcal{M}_t\neq\varnothing\), let \(m_t\) be a node in
\(\mathcal{M}_t\) with the largest information-set cardinality. We then define the
acceptable auxiliary anchors as:
\begin{equation}
\mathcal{A}_t^{\mathrm{aux}}
=
\left\{
v\in\mathcal{R}_{x,\mathrm{aux}}^+
\;\middle|\;
\operatorname{par}(v)=m_t,\;
|U(v)|=|U_t^s|
\right\},
\end{equation}
where \(\operatorname{par}(v)\) denotes the reachable teacher main node
from which the auxiliary node \(v\) is reached. Thus, the projection
compares only the number of useless operations performed after \(m_t\),
without requiring the unsuccessful tools to be identical. If
\(\mathcal{M}_t=\varnothing\), then
\(\mathcal{A}_t^{\mathrm{aux}}=\varnothing\).
Let \(\mathcal{A}_t\) be
\(\mathcal{A}_t^{\mathrm{main}}\) or
\(\mathcal{A}_t^{\mathrm{aux}}\) according to the type of \(s_t\), and
define
\begin{equation}
\rho_t
=
\mathbb{I}\!\left[
\mathcal{A}_t\neq\varnothing
\right].
\end{equation}
If \(\rho_t=1\), the student state corresponds to at least one reachable
teacher node of the same type. We formally denote the selected valid teacher anchor as \(\pi(s_t)\): for a main node, \(\pi(s_t)\) is chosen from \(\mathcal{A}_t^{\mathrm{main}}\) as a node with the maximal information set; for an auxiliary node, \(\pi(s_t)\) can be any anchor in \(\mathcal{A}_t^{\mathrm{aux}}\). If \(\rho_t=0\), no valid same-type anchor exists and the student state is treated as outside the success-reachable region.

\textbf{Critical Topological Breakpoint (CTB).}
Since the initial state is shared by the teacher and student, \(\rho_0=1\).
The CTB is defined as the first transition from a projectable to a
non-projectable student state:

\begin{equation}
\begin{aligned}
t_{\mathrm{C}} &= \min\{t:\rho_{t-1}=1,\ \rho_t=0\},\\
a_{\mathrm{C}} &= \pi(s_{t_{\mathrm{C}}-1}),
\end{aligned}
\label{eq:critical_breakpoint}
\end{equation}
where \(\pi(\cdot)\) denotes the projection mapping that assigns a valid student node to its teacher anchor. Thus, \(t_{\mathrm{C}}\) identifies the first departure from the
teacher-supported region, and the anchor \(a_{\mathrm{C}}\) is the last valid
teacher projection before that departure. This boundary captures the
earliest point at which the student either acquires an unsupported
information combination or exceeds the teacher-supported exploration
count under the corresponding main node. If the student remains projectable until the
end of the rollout but still fails, we instead view the terminal state
as the correction boundary and use its latest valid projection as the
recovery anchor.

\subsection{CTB-Guided Localized Supervision}
\label{sec:localized_distillation}
\textbf{Privileged-context Retrieval.}
We randomly sample successful and failed teacher traces from the teacher graph as privileged references, denoted as \(\mathcal{C}^{\mathrm{priv}}\).
The reference traces are not directly concatenated with the student
prefix, since the two trajectories may contain different calls and
observations. Instead, an augmented generator is conditioned on the
task, the student prefix retained before the CTB, and the
privileged references:

\begin{equation}
    c_{t_{\mathrm{C}}}^*
    =
    \operatorname{AugGen}
    \bigl(
        x,\tau_{<t_{\mathrm{C}}}^s,\mathcal{C}^{\mathrm{priv}}
    \bigr).
    \label{eq:privileged_context_generation}
\end{equation}
The training trajectory is formed by appending \(c_{t_{\mathrm{C}}}^*\) to the retained student prefix. The trajectory is then tokenized, and CTB-localized supervision is applied to the resulting training sequence. Thus, the realized student context is preserved, while the sampled teacher traces provide privileged information for generating the subsequent continuation.

\textbf{CTB-Localized Supervision.}
Let \(\widetilde{y}=(\widetilde{y}_1,\ldots,\widetilde{y}_L)\) denote the
tokenized training trajectory. We train with the masked causal language
modeling objective:
\begin{equation}
\mathcal{L}_{\mathrm{DART}}
=
-\sum_{i=1}^{L}
m_i
\log p_\theta
\bigl(
\widetilde{y}_i
\mid
\widetilde{y}_{<i},x
\bigr).
\end{equation}
Although the loss is computed at the token level, the mask is defined
over response steps: \(m_i=1\) only if token \(\widetilde{y}_i\) belongs
to an assistant response step generated after the CTB and before the
final-answer step. Tokens in the retained student prefix, user messages,
tool observations, and the final-answer step receive zero weight. Thus,
supervision is restricted to post-CTB assistant response steps, while the
pre-CTB behavior and final answer are not directly optimized.

\subsection{Progressive Self-Distillation Paradigm}
The task-specific ISTG is maintained throughout the self-distillation. At each iteration, the current student produces new rollouts, which are mapped into the shared interaction-state space and processed by the type-specific projection and breakpoint-localization procedure. Failed rollouts yield breakpoint-localized privileged-context distillation instances. As the student improves, its states remain projectable for longer portions of the interaction trajectory. The detected CTBs therefore track the evolving boundary at which the current policy first departs from the teacher-supported region. DART-SD consequently induces a self-paced curriculum that progressively shifts supervision toward recovery behaviors beyond the current tool-use capability.

\begin{table*}[t]
    \centering
    \resizebox{1.0\textwidth}{!}{
    \begin{tabular}{l|ccc|c|c|c|ccc|c}
    \specialrule{0.8pt}{0pt}{0pt}
    \multirow{2.5}{*}{\textbf{Methods}}
    & \multicolumn{3}{c|}{\textbf{FTRL}}
    & \multicolumn{1}{c|}{\textbf{BFCL}}
    & \multicolumn{1}{c|}{\textbf{ToolHop}}
    & \multicolumn{1}{c|}{$\tau$\textbf{-bench}}
    & \multicolumn{3}{c|}{\textbf{RoTBench}}
    & \multirow{2.5}{*}{\textbf{Avg.}} \\
    \cline{2-10}
    & \textbf{Solve-P}
    & \textbf{Solve-R}
    & \textbf{Solve-F1}
    & \textbf{Multi-Turn}
    & \textbf{AC}
    & \textbf{Pass\textasciicircum1}
    & \textbf{TS}
    & \textbf{PI}
    & \textbf{CF}
    & \\
    
        \cdashline{1-11}
    \rowcolor{sectiongray}
    \multicolumn{11}{c}{\textit{\textbf{Qwen3-4B}}} \\
    \cdashline{1-11}

        Base
        & 21.00
        & 26.54
        & 21.81
        & 10.14
        & 20.20
        & 15.15
        & 69.52
        & 26.31
        & 16.07
        & 25.19
        \\

        $\spadesuit$ SFT
        & 34.26
        & \cellcolor{topcolor}\textbf{49.02}
        & \cellcolor{secondcolor}\underline{37.96}
        & \cellcolor{secondcolor}\underline{14.57}
        & \cellcolor{secondcolor}\underline{40.50}
        & \cellcolor{secondcolor}\underline{21.82}
        & \cellcolor{secondcolor}\underline{71.67}
        & \cellcolor{topcolor}\textbf{44.40}
        & \cellcolor{secondcolor}\underline{24.40}
        & \cellcolor{secondcolor}\underline{37.62}
        \\

        $\spadesuit$ SCoRe-SFT
        & 23.79
        & 28.20
        & 24.61
        & 12.75
        & 25.83
        & 11.52
        & 60.36
        & 31.19
        & 18.33
        & 26.29
        \\

        $\spadesuit$ OPSD
        & 22.37
        & 29.21
        & 23.50
        & 11.00
        & 20.20
        & \cellcolor{topcolor}\textbf{23.03}
        & 68.93
        & 35.24
        & 21.67
        & 28.35
        \\


        $\diamondsuit$ FTRL-GRPO
        & \cellcolor{topcolor}\textbf{36.83}
        & 41.71
        & 37.84
        & 13.50
        & 29.25
        & 20.61
        & 70.36
        & 32.62
        & 20.24
        & 33.66
        \\

        $\diamondsuit$ ToolRL
        & 26.89
        & 33.23
        & 28.47
        & 9.88
        & 20.60
        & 18.18
        & 69.64
        & 38.21
        & 24.05
        & 29.91
        \\

        $\diamondsuit$ MatchTIR (OT)
        & 23.55
        & 29.27
        & 24.95
        & 10.38
        & 26.23
        & 16.36
        & \cellcolor{topcolor}\textbf{72.02}
        & 32.74
        & 20.36
        & 28.43
        \\

        $\diamondsuit$ MatchTIR (KM)
        & 25.54
        & 31.10
        & 26.50
        & 10.00
        & 26.63
        & \cellcolor{secondcolor}\underline{21.82}
        & \cellcolor{topcolor}\textbf{72.02}
        & 33.81
        & 21.43
        & 29.87
        \\

        \hline

        $\spadesuit$ \textbf{DART-SD (Ours)}
        & \cellcolor{secondcolor}\underline{36.70}
        & \cellcolor{secondcolor}\underline{48.16}
        & \cellcolor{topcolor}\textbf{39.77}
        & \cellcolor{topcolor}\textbf{23.88}
        & \cellcolor{topcolor}\textbf{42.11}
        & \cellcolor{topcolor}\textbf{23.03}
        & \cellcolor{topcolor}\textbf{72.02}
        & \cellcolor{secondcolor}\underline{42.38}
        & \cellcolor{topcolor}\textbf{24.52}
        & \cellcolor{topcolor}\textbf{39.17}
        \\

        \cline{1-11}
    \rowcolor{sectiongray}
    \multicolumn{11}{c}{\textit{\textbf{Qwen3-8B}}} \\
    \cdashline{1-11}

        Base
        & 21.18
        & 30.71
        & 23.48
        & 18.38
        & 28.54
        & 10.13
        & 75.52
        & 36.29
        & 22.19
        & 29.60
        \\


        $\spadesuit$ SFT
        & \cellcolor{secondcolor}\underline{38.08}
        & \cellcolor{secondcolor}\underline{50.95}
        & \cellcolor{secondcolor}\underline{41.89}
        & 19.25
        & 43.52
        & \cellcolor{secondcolor}\underline{26.06}
        & \cellcolor{secondcolor}\underline{75.95}
        & 48.81
        & 30.24
        & \cellcolor{secondcolor}\underline{41.64}
        \\

        $\spadesuit$ SCoRe-SFT
        & 31.28
        & 34.48
        & 31.58
        & 19.25
        & 30.25
        & 18.18
        & 70.36
        & 40.48
        & 24.52
        & 33.38
        \\

        $\spadesuit$ OPSD
        & 24.37
        & 34.88
        & 26.68
        & 20.50
        & 41.11
        & 21.21
        & 75.24
        & 43.21
        & 27.38
        & 34.95
        \\


        $\diamondsuit$ FTRL-GRPO
        & 37.66
        & 45.49
        & 40.22
        & \cellcolor{topcolor}\textbf{35.25}
        & 34.57
        & 23.03
        & \cellcolor{topcolor}\textbf{77.02}
        & 42.74
        & 27.02
        & 40.33
        \\

        $\diamondsuit$ ToolRL
        & 32.49
        & 41.07
        & 35.00
        & 20.50
        & \cellcolor{secondcolor}\underline{44.72}
        & 25.45
        & 75.71
        & \cellcolor{secondcolor}\underline{51.43}
        & \cellcolor{secondcolor}\underline{33.10}
        & 39.94
        \\

        $\diamondsuit$ MatchTIR (OT)
        & 29.45
        & 34.89
        & 30.65
        & 21.38
        & 38.79
        & 24.24
        & 75.00
        & 40.36
        & 25.36
        & 35.57
        \\

        $\diamondsuit$ MatchTIR (KM)
        & 33.07
        & 40.42
        & 35.37
        & 23.25
        & 41.71
        & \cellcolor{secondcolor}\underline{26.06}
        & 75.71
        & 42.26
        & 26.90
        & 38.31
        \\

        \hline

        $\spadesuit$ \textbf{DART-SD (Ours)}
        & \cellcolor{topcolor}\textbf{42.00}
        & \cellcolor{topcolor}\textbf{54.13}
        & \cellcolor{topcolor}\textbf{45.66}
        & \cellcolor{secondcolor}\underline{27.63}
        & \cellcolor{topcolor}\textbf{45.03}
        & \cellcolor{topcolor}\textbf{27.12}
        & 75.83
        & \cellcolor{topcolor}\textbf{57.38}
        & \cellcolor{topcolor}\textbf{35.48}
        & \cellcolor{topcolor}\textbf{45.58}
        \\

        \specialrule{0.8pt}{0pt}{0pt}
    \end{tabular}
    }

    \caption{
    Performance comparison of different training methods on five tool-use
    benchmarks using Qwen3-4B and Qwen3-8B backbones.
    Training-based baselines are organized into distillation-based $\spadesuit$ and
    reinforcement learning $\diamondsuit$ paradigms.
    All trainable methods are trained on FTRL and evaluated on both the
    in-domain FTRL test set and four out-of-domain benchmarks.
    The \sethlcolor{topcolor}\hl{\textbf{best}} and
    \sethlcolor{secondcolor}\hl{\mbox{\underline{second-best}}}
    results within each backbone are highlighted.
    }
    \label{tab:main}
\end{table*}

\section{Experiments}
\subsection{Experimental Setup}

\textbf{Datasets.}
We train the models on the \textbf{FTRL} dataset \cite{ye2025feedback}, which comprises over 2,000 automatically constructed tool-use environments with verifiable feedback. FTRL covers four types of task structures: Single, containing a single sub-question; Para-Single, consisting of multiple independent sub-questions executable in parallel; Multi, comprising a sequence of dependent sub-questions; and Para-Multi, combining both independent and dependent sub-questions. We evaluate DART-SD on five tool-use benchmarks, including FTRL as the in-domain test set, and \textbf{BFCL} \cite{patil2025berkeley}, \textbf{ToolHop} \cite{ye2025toolhop}, \textbf{$\tau$-bench} \cite{yao2024tau}, and \textbf{RoTBench} \cite{ye2024rotbench} as out-of-domain benchmarks to assess generalization. 

\begin{figure}[h]
\centering
\includegraphics[]{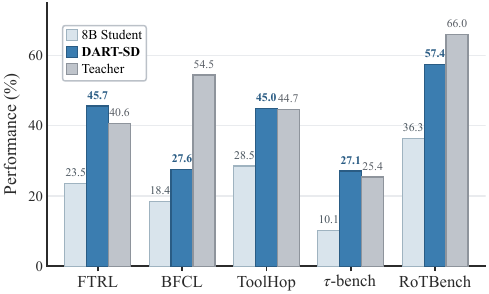}
\caption{Performance comparison of Qwen3-8B, DART-SD, and the teacher across five tool-use benchmarks. DART-SD improves upon Qwen3-8B on all benchmarks and \textbf{surpasses} the teacher on FTRL, ToolHop, and $\tau$-bench.}
\label{fig:comparison}
\end{figure}

\textbf{Baselines.}
We evaluate DART-SD using both Qwen3-4B and Qwen3-8B backbones \cite{qwen3technicalreport}. For each backbone, we first report the performance of the pretrained model without task-specific adaptation, denoted as \textit{Base}. 
The remaining baselines are grouped into two optimization paradigms. \textbf{(1) Distillation-based Paradigms:} standard \textit{SFT}, \textit{SCoRe-SFT} \cite{score2026}, and \textit{OPSD} \cite{opsd2026}. \textbf{(2) RL Paradigms:} \textit{FTRL-GRPO} \cite{ye2025feedback}, \textit{ToolRL} \cite{qian2026toolrl},  and \textit{MatchTIR} \cite{qu2026matchtir} (evaluated using both \textit{MatchTIR-OT} and \textit{MatchTIR-KM}). To ensure a strictly controlled comparison, all methods are trained and evaluated under a no-thinking configuration.

\textbf{Evaluation Metrics.}
To ensure a rigorous and standardized assessment, we strictly adhere to the official evaluation protocols for each respective benchmark. \textbf{(1) FTRL:} We evaluate trajectory precision and task completeness using \textit{Solve-P} and \textit{Solve-R} along with their harmonic mean, \textit{Solve-F1}. 
\textbf{(2) BFCL:} We report the average \textit{Multi-Turn} score over Base, Miss Function, Miss Parameter and Long Context.
\textbf{(3) ToolHop:} We evaluate the model’s ability to resolve complex multi-hop tool dependencies using \textit{Answer Correctness (AC)}. \textbf{(4) $\tau$-bench:} We assess performance on real-world API orchestration tasks using \textit{Pass\textasciicircum1}. \textbf{(5) RoTBench:} We evaluate fine-grained execution robustness through \textit{Tool Selection (TS)}, \textit{Parameter Identification (PI)}, and \textit{Content Filling (CF)}.

\textbf{Implementation Details.}
We initialize the student model from Qwen3-4B and Qwen3-8B. Teacher trajectories are collected from a mixed pool of Qwen3.6-27B~\cite{qwen3.6-27b} and GLM-5.2~\cite{zeng2026glm}. We perform five iterations of CTB-guided localized SFT over all 2,215 tasks in the FTRL training set. At each iteration, the current student checkpoint generates eight trajectories per task with a temperature of 0.7, a maximum generation length of 4,096 tokens, and at most nine interaction turns. The model is then fine-tuned for one epoch using breakpoint-localized loss masking, with a learning rate of $5\times10^{-7}$ and a batch size of 32. Each augmented training context contains two positive references and one negative reference, and each positive reference is accompanied by a teacher-generated analysis.

\subsection{Experimental Results}

\textbf{Overall Performance.}
As shown in Table~\ref{tab:main}, DART-SD demonstrates robust and superior performance on both the in-domain FTRL dataset and four out-of-domain tool-use benchmarks. Compared with distillation-based methods and reinforcement learning methods, DART-SD delivers consistent empirical gains and achieves the strongest overall performance. This robust cross-domain superiority confirms that our topology-aware localized tuning learns generalizable and transferable tool-use capabilities, rather than merely overfitting to the FTRL training distribution.
Crucially, DART-SD maintains strong performance across different parameter scales. It achieves the highest average performance with both the Qwen3-8B and Qwen3-4B backbones, while obtaining the best or competitive results on most individual metrics. These results indicate that the effectiveness of topology-aware localized tuning is not limited to a particular model size and remains robust when applied to a smaller backbone.

\textbf{Effectiveness of Capability Distillation.}
Figure~\ref{fig:comparison} compares the performance of the base Qwen3-8B student, DART-SD, and the teacher across five tool-use benchmarks. For visualization, we report Solve-F1 on FTRL, the average Multi-Turn score on BFCL, AC on ToolHop, Pass\textasciicircum1 on $\tau$-bench, and PI on RoTBench. Initialized from the same 8B base model, DART-SD delivers substantial improvements across all five benchmarks. Notably, the resulting student surpasses the teacher on FTRL, ToolHop, and $\tau$-bench, despite using substantially fewer parameters. 
These results suggest that DART-SD goes beyond direct trajectory imitation. By leveraging topology-aware recovery paths, it effectively extracts and generalizes reusable tool-use behaviors from the structural knowledge encoded in the ISTG, enabling the smaller student to outperform its teacher on several benchmarks.

\definecolor{goldbg}{RGB}{242,242,242}

\begin{table}[h]
\centering
\begin{tabular}{l|ccccc|c}
\specialrule{0.8pt}{0pt}{0pt}
\multirow{2}{*}{\textbf{Task}}
& \multicolumn{5}{c|}{\textbf{DART-SD}}
& \multirow{2}{*}{\textbf{Golden}} \\

\cline{2-6}

& Iter1 & Iter2 & Iter3 & Iter4 & Iter5 & \\

\hline

Single
& 1.23 & 1.15 & 1.10 & 1.09
& \cellcolor{secondcolor}1.07
& \cellcolor{goldbg}1.00 \\

Multi
& 3.83 & 4.63 & 4.54 & 4.49
& \cellcolor{secondcolor}4.41
& \cellcolor{goldbg}4.71 \\

Para-Single
& 3.29 & 2.65 & 2.27 & 2.26
& \cellcolor{secondcolor}2.22
& \cellcolor{goldbg}2.11 \\

Para-Multi
& 7.25 & 6.48 & 5.97 & 5.82
& \cellcolor{secondcolor}5.62
& \cellcolor{goldbg}6.97 \\

\hline

Overall
& 4.23 & 3.99 & 3.71 & 3.65
& \cellcolor{topcolor}\textbf{3.55}
& \cellcolor{goldbg}\textbf{4.02} \\

\hline

Solve-F1
& 40.37 & 42.95 & 43.78 & 44.67
& \textbf{45.66}
& -- \\

\specialrule{0.8pt}{0pt}{0pt}
\end{tabular}
\caption{Average tool-call length of successful trajectories across progressive SFT iterations on the FTRL test set. DART-SD progressively improves Solve-F1 while shortening its tool traces, eventually producing more efficient traces than the golden solutions given during data construction.}
\label{tab:ablation_better_trace}
\end{table}

\textbf{Tool-Call Length of Successful Trajectories.}
To verify whether DART-SD improves reasoning rather than memorizing references, we track the evolution of successful trajectories across iterations. We measure task performance (\textit{Solve-F1}) and efficiency (average tool calls) on the FTRL test set, and compare the final model against the golden references. The golden reference denotes the tool trajectory provided by the FTRL construction pipeline.
As shown in Table~\ref{tab:ablation_better_trace}, \textit{Solve-F1} steadily improves while trajectory length decreases. The average number of tool calls drops from 4.23 at Iter1 to 3.55 at Iter5. This confirms that the model learns \textbf{more efficient tool-use strategies} instead of relying on brute-force exploration or longer execution chains. This reduction is especially clear in complex multi-step and parallel tasks, suggesting that our method effectively removes redundant tool calls while maintaining high success rates.
Remarkably, the trajectories of our final model are even \textbf{shorter than the golden references} (3.55 vs.\ 4.02). This highlights that DART-SD successfully discovers optimized shortcuts rather than blindly following the provided subtask structures.

\definecolor{c1}{RGB}{235,246,251}
\definecolor{c2}{RGB}{218,237,247}
\definecolor{c3}{RGB}{198,226,241}
\definecolor{c4}{RGB}{174,213,233}
\definecolor{c5}{RGB}{145,196,223}

\begin{table}[h]
\centering
\begin{tabular}{l|ccccc}
\specialrule{0.8pt}{0pt}{0pt}

\multirow{2}{*}{\textbf{Task}}
& \multicolumn{5}{c}{\textbf{CTB Position}} \\

\cline{2-6}

& Iter1 & Iter2 & Iter3 & Iter4 & Iter5 \\

\hline

Single
& 0.034 & 0.186 & 0.137 & 0.249 & 0.077 \\

Multi
& 0.500 & 1.624 & 1.729 & 1.876 & 1.953 \\

Para-Single
& 0.095 & 0.219 & 0.299 & 0.328 & 0.313 \\

Para-Multi
& 0.395 & 1.524 & 1.692 & 1.791 & 1.816 \\

\hline

Overall
& \cellcolor{c1}0.348
& \cellcolor{c2}1.185
& \cellcolor{c3}1.310
& \cellcolor{c4}1.421
& \cellcolor{c5}1.452 \\

\hline

$\Delta$ vs. Iter1
& --
& +0.837
& +0.962
& +1.073
& \textbf{+1.104} \\

\specialrule{0.8pt}{0pt}{0pt}
\end{tabular}
\caption{
Average CTB positions of failed training trajectories across progressive SFT.
Larger values indicate that the first departure from empirically recoverable
behavior occurs later, meaning that the model correctly executes a longer
trajectory prefix before localized recovery is required.
}
\label{tab:ablation_CTB}
\end{table}

\textbf{CTB Position Shifts within Failed Trajectories.}
To analyze how the student’s capability boundary expands during progressive training, we track the average CTB position within failed trajectories across iterations. A later CTB position indicates that the model remains within the empirically recoverable region for a longer prefix before its first departure. As shown in Table~\ref{tab:ablation_CTB}, the average CTB position steadily advances from 0.348 at Iter1 to 1.452 at Iter5, indicating that the first departure from recoverable behavior is progressively delayed. This trend is particularly evident in complex multi-hop and parallel multi-hop tasks, where increasingly longer valid prefixes are learned. These results validate the effectiveness of CTB-guided localized supervision. By identifying the current capability boundary, DART-SD performs teacher-guided resampling from the first departure point and supervises only the recovery suffix while preserving already-mastered prefixes. As the boundary moves deeper across iterations, the student progressively acquires more complex tool-use behaviors, demonstrating the effectiveness of DART-SD for long-horizon tasks.

\begin{table}[h]
\centering
\begin{tabular}{l|ccc}
\specialrule{0.8pt}{0pt}{0pt}
\textbf{Method}
& \textbf{FTRL}
& \textbf{BFCL}
& \textbf{ToolHop} \\

\hline

Qwen3-8B
& 29.74
& 40.00
& 42.21 \\

FTRL-GRPO
& 32.85
& 41.50
& 36.72 \\

ToolRL
& 26.72
& 34.25
& 32.93 \\

MatchTIR (KM)
& \cellcolor{secondcolor}\underline{37.33}
& \cellcolor{secondcolor}\underline{47.13}
& \cellcolor{secondcolor}\underline{46.16} \\

\hline

\textbf{DART-SD}
& \cellcolor{topcolor}\textbf{41.03}
& \cellcolor{topcolor}\textbf{49.75}
& \cellcolor{topcolor}\textbf{46.43} \\

\specialrule{0.8pt}{0pt}{0pt}
\end{tabular}
\caption{Performance comparison under the thinking setting. Results are reported using FTRL Solve-F1, the average BFCL Multi-Turn score, and ToolHop AC.}
\label{tab:ablation_think}
\end{table}

\textbf{Effect of Thinking Mode.}
Table~\ref{tab:ablation_think} evaluates DART-SD under the thinking setting. All trainable baselines are trained with thinking enabled, whereas the Qwen3-8B base model enables thinking only at inference time. For DART-SD, teacher trajectories are generated in thinking mode, and both the teacher's reasoning traces and tool-call trajectories are incorporated into the resampling context for student training.
The effect of explicit thinking varies across methods and benchmarks. To provide a fair and controlled comparison, we therefore use the no-thinking configuration in all other experiments unless otherwise specified, allowing us to more directly evaluate the proposed training framework. Importantly, DART-SD still achieves the best performance across all three benchmarks under the thinking setting, demonstrating that CTB-guided localized supervision remains effective when explicit thinking is enabled.

\begin{table}[h]
\centering
\begin{tabular}{l|ccccc}
\specialrule{0.8pt}{0pt}{0pt}

\textbf{Method}
& \textbf{IFEval}
& \textbf{AIME24}
& \textbf{AIME25}
& \textbf{MMLU}
& \textbf{Avg.} \\

\hline

Qwen3-8B
& 34.75
& 46.67
& 23.33
& 70.94
& 43.92 \\

SFT
& 35.30
& 43.33
& 26.67
& 71.43
& 44.18 \\

\textbf{DART-SD}
& \cellcolor{topcolor}\textbf{45.29}
& \cellcolor{topcolor}\textbf{50.00}
& \cellcolor{topcolor}\textbf{30.00}
& \cellcolor{topcolor}\textbf{74.27}
& \cellcolor{topcolor}\textbf{49.89} \\

\specialrule{0.8pt}{0pt}{0pt}
\end{tabular}
\caption{
General capability evaluation on representative benchmarks.
The \sethlcolor{topcolor}\hl{\textbf{best}} results are highlighted.
}
\label{tab:general}
\end{table}

\textbf{General Capability Preservation.}
To evaluate whether DART-SD preserves the model’s general capabilities beyond tool-use tasks, we further evaluate it under a thinking-enabled setting on four standard benchmarks: IFEval \cite{zhou2023IFEval} for instruction following, AIME24 and AIME25 for mathematical reasoning, and MMLU \cite{hendrycks2020mmlu} for general knowledge. We report strict prompt-level accuracy on IFEval, pass@10 on AIME24 and AIME25, and accuracy on MMLU. As shown in Table~\ref{tab:general}, DART-SD consistently outperforms both the pretrained model and standard SFT across all benchmarks, improving the average score from 43.92 to 49.89. These results suggest that topology-aware localized supervision not only enhances tool-use ability but also preserves the model’s general reasoning and instruction-following capabilities.


\begin{table}[h]
\centering
\begin{tabular}{l|ccc}
\specialrule{0.8pt}{0pt}{0pt}

\textbf{Method}
& \textbf{Solve-P}
& \textbf{Solve-R}
& \textbf{Solve-F1} \\

\hline

Qwen3-8B
& 21.18
& 30.71
& 23.48 \\

+SD
& 36.23
& 44.46
& 38.10 \\

+CTB
& 36.62
& 46.32
& 39.51 \\

+Progressive SFT
& 41.32
& 49.65
& 43.93 \\

+ISTG (\textbf{Ours})
& \cellcolor{topcolor}\textbf{42.00}
& \cellcolor{topcolor}\textbf{54.13}
& \cellcolor{topcolor}\textbf{45.66} \\

\specialrule{0.8pt}{0pt}{0pt}
\end{tabular}
\caption{
Component ablation of DART-SD on the FTRL test set using Solve-P, Solve-R, and Solve-F1.
}
\label{tab:ablation_component}
\end{table}

\textbf{Ablation Study on Different Components.}
Table~\ref{tab:ablation_component} presents the component ablation of DART-SD to evaluate the contribution of each core design.
First, applying Self-Distillation (SD) directly over the base model yields substantial gains, demonstrating the efficacy of learning from the agent's own exploratory trajectories.
Second, introducing CTB-guided localized supervision further improves performance by restricting the loss calculation exclusively to the post-breakpoint recovery suffix, thereby avoiding direct global supervision on already-mastered prefixes.
Third, Progressive Self-Distillation introduces iterative rollout and localized correction rounds, confirming the necessity of a continuous iterative adaptation paradigm over one-shot training.
Finally, ISTG replaces LLM-judge-based breakpoint detection with topology-aware state projection and structured recovery, enabling more accurate CTB identification and structurally compatible supervision. Together, these complementary components yield the strongest performance.

\section{Conclusion}

In this paper, we propose \textbf{DART-SD}, a topology-aware self-distillation framework for multi-turn tool-calling agents. DART-SD models execution as an \textit{Interaction-State Transition Graph} (\textbf{ISTG}), revealing the diamond topology induced by order-independent valid exploration. Building on the ISTG, DART-SD identifies the \textit{Critical Topological Breakpoint} (\textbf{CTB}), retrieves recovery references from the ISTG, and performs progressive self-distillation via CTB-guided localized supervision while preserving the valid interaction prefix.
Extensive experiments demonstrate that DART-SD consistently outperforms both distillation and reinforcement learning baselines across multiple in-domain and out-of-domain tool-use benchmarks. Beyond improving task success, DART-SD enables more efficient tool-use behaviors by preserving exploration, reducing redundant tool calls, and progressively mastering more complex tool-use behaviors throughout self-distillation.
More broadly, our findings suggest that effective agent distillation should be guided by interaction-state topology rather than rigid trajectory imitation. We hope this perspective encourages future research on structure-aware agent training and graph representations for long-horizon reasoning and decision-making.

\clearpage

\bibliographystyle{plainnat}
\bibliography{main}

\end{document}